\pdfoutput=1
\documentclass[letterpaper]{article} %
\usepackage{aaai23}  %
\usepackage{times}  %
\usepackage{helvet}  %
\usepackage{courier}  %
\usepackage[hyphens]{url}  %
\usepackage{graphicx} %
\def\UrlFont{\rm}  %
\usepackage{natbib}  %
\usepackage{caption} %
\usepackage{algorithm}
\usepackage{algorithmic}

\usepackage{newfloat}
\usepackage{listings}
\DeclareCaptionStyle{ruled}{labelfont=normalfont,labelsep=colon,strut=off} %
\floatstyle{ruled}
\newfloat{listing}{tb}{lst}{}
\floatname{listing}{Listing}
\usepackage{breakcites}
\usepackage[hyperindex,breaklinks=true]{hyperref} %

\usepackage{epsfig}
\usepackage{graphicx}
\usepackage{amsmath}
\usepackage{amssymb}
\usepackage{bm}

\usepackage{bbm}
\usepackage{subcaption}
\usepackage{multirow}
\usepackage{booktabs}   %
\usepackage{makecell}   %
\usepackage{changepage} %
\usepackage{comment}

\usepackage{wrapfig}

\usepackage{pifont}
\newcommand{\cmark}{\ding{51}}
\newcommand{\xmark}{\ding{55}}

\usepackage{tabularx}
\newcolumntype{R}{>{\raggedleft\arraybackslash}X}

\usepackage{longtable}

\usepackage{inconsolata}

\makeatletter
\usepackage{xspace}
\def\@onedot{\ifx\@let@token.\else.\null\fi\xspace}
\DeclareRobustCommand\onedot{\futurelet\@let@token\@onedot}

\def\eg{\emph{e.g}\onedot} 
\def\ie{\emph{i.e}\onedot} \def\Ie{\emph{I.e}\onedot}
 
 \def\vs{\emph{vs}\onedot}

\newcommand{\figref}[1]{Fig\onedot~\ref{#1}}
\newcommand{\equref}[1]{Eq\onedot~\eqref{#1}}
\newcommand{\secref}[1]{Sec\onedot~\ref{#1}}
\newcommand{\tabref}[1]{Tab\onedot~\ref{#1}}

\usepackage{csquotes}

\usepackage[normalem]{ulem}

\usepackage{multirow}

\usepackage{listings}

\usepackage{adjustbox}

\usepackage{fancybox, graphicx}

\usepackage{threeparttable}

\usepackage{color, colortbl}
\definecolor{Gray}{gray}{0.9}
\newcolumntype{g}{>{\columncolor{Gray}}r}
\definecolor{highlightRowColor}{rgb}{0.95, 0.95, 1}

\newcommand*{\belowrulesepcolor}[1]{%
  \noalign{%
    \kern-\belowrulesep 
    \begingroup 
      \color{#1}%
      \hrule height\belowrulesep 
    \endgroup 
  }%
} 
\newcommand*{\aboverulesepcolor}[1]{%
  \noalign{%
    \begingroup 
      \color{#1}%
      \hrule height\aboverulesep 
    \endgroup 
    \kern-\aboverulesep 
  }%
} 

\newcommand{\beginsupplement}{
    \setcounter{table}{0}
    \renewcommand{\thetable}{S\arabic{table}}%
    \setcounter{figure}{0}
    \renewcommand{\thefigure}{S\arabic{figure}}%
    \setcounter{equation}{0}
    \renewcommand{\theequation}{S\arabic{equation}}
}

\usepackage[resetlabels]{multibib}
\newcites{supp}{References}

\usepackage{url}            %
\usepackage{booktabs}       %
\usepackage{amsfonts}       %
\usepackage{nicefrac}       %
\usepackage{microtype}      %
\usepackage[dvipsnames,table,xcdraw]{xcolor}
\definecolor{airforceblue}{rgb}{0.36, 0.54, 0.66}

\definecolor{light-gray}{gray}{0.7}
\newcommand{\gcmark}{\textcolor{light-gray}{\cmark}}

\title{Occupancy Planes for Single-view RGB-D Human Reconstruction}
\author{
    Xiaoming Zhao, Yuan-Ting Hu, Zhongzheng Ren, Alexander G. Schwing
}
\affiliations{

    University of Illinois Urbana-Champaign \\
    \url{https://github.com/Xiaoming-Zhao/oplanes}
}

\begin{document}

\maketitle

\begin{abstract}
Single-view RGB-D human reconstruction with implicit functions is often formulated as per-point classification. Specifically,  a set of 3D locations within the view-frustum of the camera are first projected independently onto the image and a corresponding feature is subsequently extracted for each 3D location. The feature of each 3D location is then used to classify independently whether the corresponding 3D point is inside or outside the observed object. This procedure leads to sub-optimal results because correlations between predictions for neighboring locations are only taken into account implicitly via the extracted features. For more accurate results we propose the \emph{occupancy planes} (OPlanes) representation, which enables to formulate single-view RGB-D human reconstruction as occupancy prediction on planes which slice through the camera's view frustum. Such a representation provides more flexibility than voxel grids and enables to better leverage correlations than per-point classification. On the challenging S3D data we observe a simple classifier based on the OPlanes representation to yield compelling results, especially in difficult situations with partial occlusions due to other objects and partial visibility, which haven't been addressed by prior work.
\end{abstract}

\vspace{-0.1cm}
\section{Introduction}
Reconstructing the 3D shape of humans~\cite{guan2009estimating,tong2012scanning,bogo2016keep,lassner2017unite,guler2019holopose,xiang2019monocular,yu2017bodyfusion,varol2018bodynet,Zheng2019DeepHuman3H,Saito2019PIFuPI,renredo2021} has attracted extensive attention. It enables numerous applications such as AR/VR content creation, virtual try-on in the fashion industry, and image/video editing. In this paper, we focus on the task of human reconstruction from a single RGB-D image and its camera information. A single-view setup is simple and alleviates the tedious capturing of  sequences from multiple locations. However, while capturing of data is simplified, the task is more challenging  because of the ambiguity in inferring the invisible part of the 3D human shape given only the camera facing part.

 \begin{figure}[!t]
\centering
\captionsetup[subfigure]{width=\columnwidth}
\centering
\includegraphics[width=0.85\columnwidth]{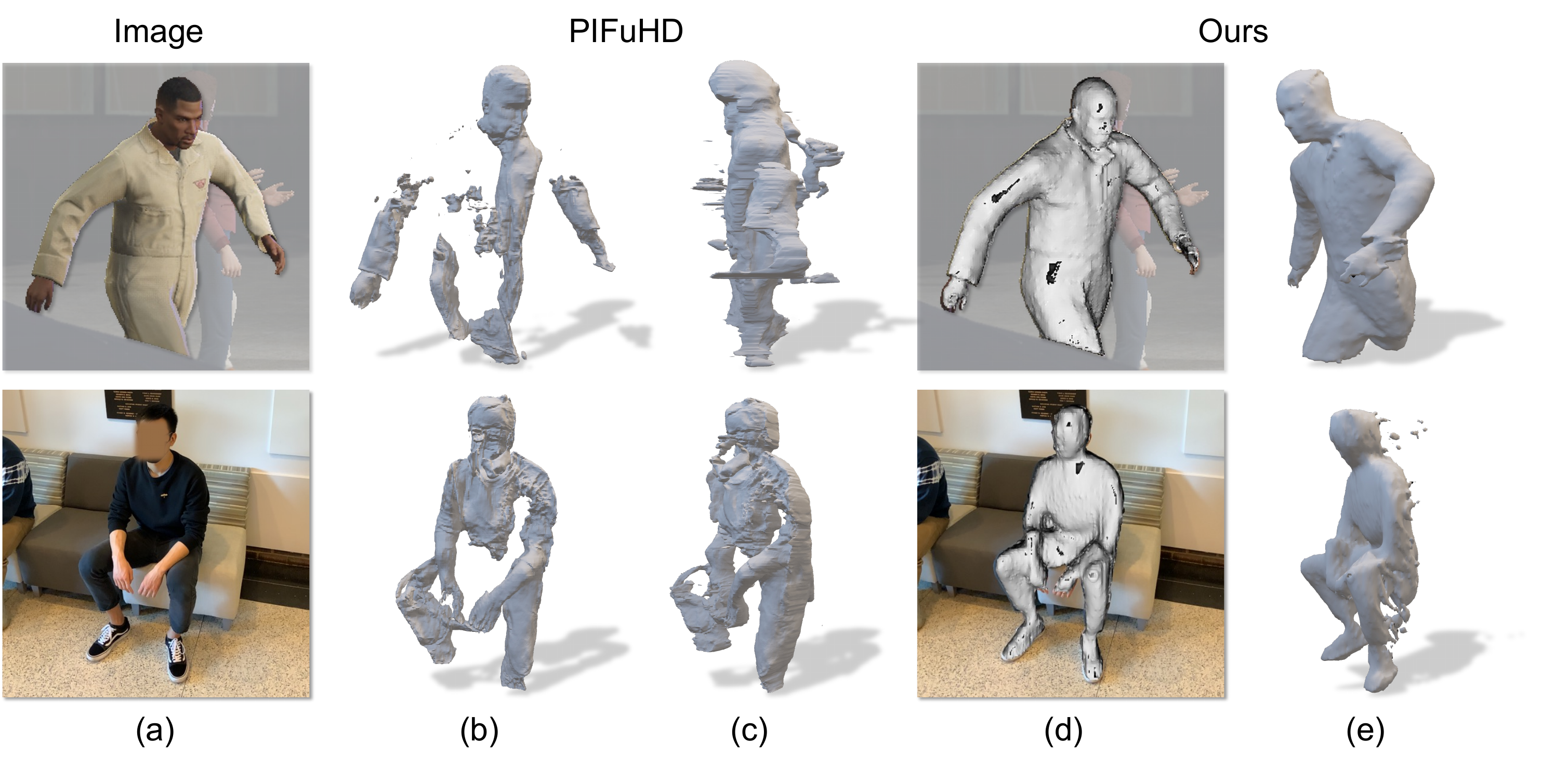} 
\captionsetup{width=\columnwidth}
\vspace{-0.4cm}
\caption{Reconstruction results compared to PIFuHD~\cite{Saito2020PIFuHDMP} on S3D~\cite{Hu2021SAILVOS3A} data ($1^\text{st}$ row) and real-world data ($2^\text{nd}$ row).
\textbf{(a)}: input image;
\textbf{(b)} and \textbf{(c)}:  results from PIFuHD. It struggles with partial visibility and non-standing poses;
\textbf{(d)}: our reconstruction overlaying  the input image with perspective camera projection.
\textbf{(e)}: another view of our reconstruction.
}
\label{fig: teaser}
\vspace{-0.6cm}
\end{figure}

A promising direction for 3D reconstruction of a human from a single image is the use of implicit functions. Prior works which use implicit functions~\cite{Saito2019PIFuPI,Saito2020PIFuHDMP} have shown compelling results on single view human reconstruction. In common to all these prior works is the use of per-point classification. Specifically, prior works formulate single-view human reconstruction by first projecting a point onto the image. A pixel aligned feature is subsequently extracted and used to classify whether the 3D point is inside or outside the observed human.

While such a formulation is compelling, it is challenged by the fact that occupancy for every point is essentially predicted independently~\cite{Saito2019PIFuPI,Saito2020PIFuHDMP,Mescheder2019OccupancyNL}. While image features permit to take context into account implicitly, the point-wise prediction often remains noisy. This is particularly true if the data is more challenging than the one used in prior works. For instance, we observe prior works to struggle if humans are only partially visible,~\ie, if we are interested in only reconstructing the observed part of a partially occluded person.

To address this concern, we propose to formulate single-view RGB-D human reconstruction as an occupancy-plane-prediction task. For this, we introduce the novel \emph{occupancy planes} (OPlanes) representation. The OPlanes representation consists of multiple  image-like planes which slice in a fronto-parallel manner through the camera's view frustum and indicate at every pixel location  the occupancy for the corresponding 3D point. Importantly, the OPlanes representation permits to adaptively adjust the number and location of the occupancy planes  during inference and training. Therefore, its resolution is more flexible than that of a classical voxel grid representation~\cite{varol2018bodynet,maturana2015voxnet}. Moreover, the plane structure naturally enables the model to benefit more directly from correlations between  neighboring locations within a plane than unstructured representations like point clouds~\cite{Fan2017APS,qi2017pointnet} and implicit representations with per-point queries~\cite{Saito2019PIFuPI,Saito2020PIFuHDMP,Mescheder2019OccupancyNL}. 
To summarize, our contributions are two-fold: 1) we propose the  OPlanes representation for single view RGB-D human reconstruction; 2) we verify that exploiting correlations within planes  is beneficial for 3D human shape reconstruction as illustrated in \figref{fig: teaser}.

We evaluate the proposed approach on the challenging S3D~\cite{Hu2021SAILVOS3A} data and observe improvements over prior reconstruction work~\cite{Saito2020PIFuHDMP,Chibane2020ImplicitFI} by a  margin, particularly for occluded or partially visible humans. We also provide a comprehensive analysis to validate each of the  design choices and  results on real-world data.

\section{Related Work}

3D human reconstruction~\cite{guan2009estimating,tong2012scanning,yang2016estimation,zhang2017detailed,bogo2016keep,lassner2017unite,guler2019holopose,kolotouros2019convolutional,xiang2019monocular,xu2019denserac,yu2017bodyfusion,Zheng2019DeepHuman3H,varol2018bodynet} has been extensively studied for the last few decades. We first discuss the most relevant works on single-view human reconstruction~\cite{gabeur2019moulding} and group them into two categories, template-based models and non-parametric models. Then we review the common 3D representations.

\noindent\textbf{Template-based models for single-view human reconstruction.} 
Parametric human models such as SCAPE~\cite{anguelov2005scape} and SMPL~\cite{bogo2016keep} are widely used for human reconstruction. These methods~\cite{kanazawa2018end,varol2018bodynet,Zheng2019DeepHuman3H,Huang2020ARCHAR} use  the human body shape as a prior to regularize the prediction space and predict or fit the   low-dimensional parameters of a human body model. Specifically, HMR~\cite{kanazawa2018end} learns to predict the human shape by regressing the parameters of SMPL from a single image. BodyNet~\cite{varol2018bodynet} predicts a 3D voxel grid of the human shape and fits the SMPL body model to the predicted volumetric shape. DeepHuman~\cite{Zheng2019DeepHuman3H} utilizes the SMPL model as an initialization and further refines it with deep nets. Although  parametric human models are deformable and can capture various complex human body poses and different body measurements, these methods generally do not consider surface details such as hair, clothing as well as accessories.

\noindent\textbf{Non-parametric models for single-view human reconstruction.} Non-parametric methods for human reconstruction~\cite{Saito2019PIFuPI,Saito2020PIFuHDMP,He2020GeoPIFuGA,gabeur2019moulding,wang2020normalgan,renredo2021} gained popularity recently as they are more flexible in recovering surface details compared to template-based methods. Among them, using implicit function~\cite{sclaroff1991generalized} to predict human shape achieves state-of-the-art results~\cite{Saito2020PIFuHDMP}, showing that the expressivity of neural nets enables to memorize the human body shape. To achieve this, the task is usually formulated as a per-point classification,~\ie, classifying  every point in a 3D space independently into either inside or outside of the observed body. For this, PIFu~\cite{Saito2019PIFuPI} reconstructs the human shape from an image encoded into a feature map, from which it learns an implicit function to predict per-point occupancy. PIFuHD~\cite{Saito2020PIFuHDMP} employs a two level implicit predictor and  incorporates  normal information to recover high quality surfaces. GeoPIFu~\cite{He2020GeoPIFuGA} learns additionally latent voxel features to encourage shape regularization. Hybrid methods have also been studied~\cite{Huang2020ARCHAR,Cao2022JIFFJI}, combining human shape templates with a non-parametric framework. These methods usually yield reconstruction results with surface details. However, in common to all the aforementioned methods, the per-point classification formulation doesn't \emph{directly} take  correlations between neighboring 3D points  into account. Therefore, predictions remain noisy, particularly in challenging situations with occlusions or partial visibility. Because of this, prior works usually consider images where the whole human body is visible and roughly centered. In contrast, for broader applicability and more accurate results in challenging situations, we propose the OPlanes representation. 

\noindent\textbf{3D representations.} 
Various 3D representation have been developed, such as voxel grids~\cite{varol2018bodynet,maturana2015voxnet,lombardi2019neural,ren-cvpr2022-nvos}, meshes~\cite{lin2021end,wang2018pixel2mesh,wu2022casa}, point clouds~\cite{qi2017pointnet,Fan2017APS,wu2020multi,aliev2020neural}, implicit functions~\cite{Mescheder2019OccupancyNL,Saito2019PIFuPI,Saito2020PIFuHDMP,He2020GeoPIFuGA,hong2021stereopifu,peng2021neural}, layered representations~\cite{shade1998layered,zhou2018stereo,srinivasan2019pushing,tucker2020single, zhao-gmpi2022} and hierarchical representations~\cite{meagher1982geometric,hane2017hierarchical,yu2021plenoctrees}. For human body shape reconstruction,  template-based representations~\cite{anguelov2005scape,bogo2016keep,pavlakos2019expressive,osman2020star} are also popular. The proposed OPlanes representation combines the benefits of both layered and implicit representations. 
Compared to voxel grids, OPlanes is more flexible, enabling prediction at different resolutions due to its implicit formulation  of  occupancy-prediction of an entire  plane. 
Compared to unstructured representations such as implicit  functions or point clouds, OPlanes benefits from its increased context of a per-plane prediction as opposed to a per-pixel/point prediction. 
Concurrently, Fourier occupancy field (FOF)~\cite{feng2022fof} proposes to use a 2D field orthogonal to the view direction to represent the occupancy. Different from FOF, where coefficients for Fourier basis functions are estimated for each position on the 2D field, OPlanes directly regress to the occupancy value.

\section{Method}

\begin{figure*}[!t]
\centering
\captionsetup[subfigure]{width=0.8\textwidth}
    \centering
    \hspace*{-0.3cm}
    \includegraphics[width=0.75\textwidth]{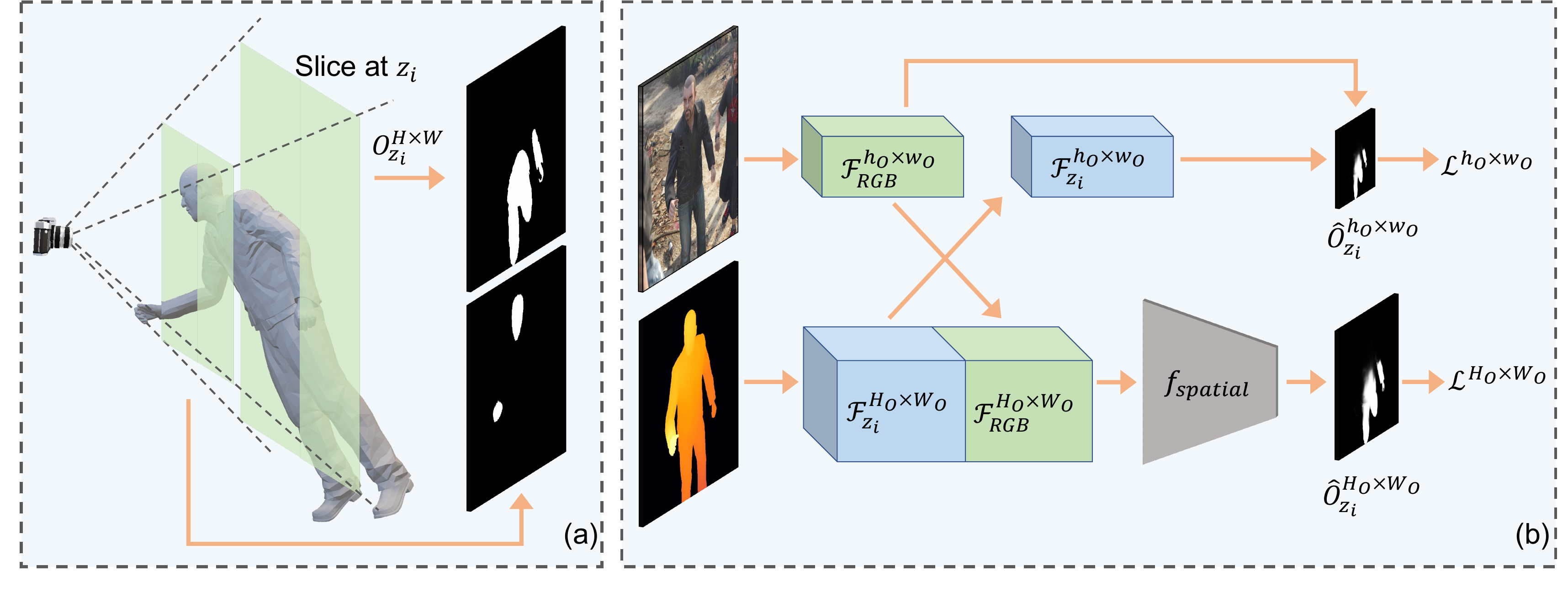} 
    \captionsetup{width=\textwidth}
    \vspace{-0.4cm}
    \caption{Occupancy planes (OPlanes) overview.
    \textbf{(a)} Occupancy plane $O_{z_i}^{H\times W}$ stores the occupancy information (black plane on the right) at a specific slice (light green plane on the left) in the view frustum. White pixels indicate ``inside'' the mesh (\secref{sec: oplane}). 
    \textbf{(b)} Given RGB-D data and a mask, our approach takes a specific depth $z_i$ as input and predicts the corresponding occupancy plane $\widehat{O}_{z_i}^{H_O\times W_O}$ (\secref{sec: oplane pred}).
    The convolutional neural network $f_\text{spatial}$ explicitly considers context information for each pixel on the occupancy plane, which we find to be beneficial.
    During training, we not only supervise $\widehat{O}_{z_i}^{H_O\times W_O}$ through loss $\mathcal{L}^{H_O \times W_O}$ but we also supervise the intermediate feature $\widehat{O}_{z_i}^{h_O\times w_O}$ with loss $\mathcal{L}^{h_O \times w_O}$ (\secref{sec: oplane train}).
    }
    \label{fig: example img}
\vspace{-0.5cm}
\end{figure*}

\subsection{Overview}\label{sec: approach overview}

Given an RGB image, a depth map, a mask highlighting the human of interest in the image as well as the intrinsic camera parameters, our goal is to reconstruct a spatially-aligned human mesh $\mathcal{M}$.
    
To generate the mesh $\mathcal{M}$, we introduce the \textit{Occupancy Planes} (OPlanes) representation, a plane-based  representation of the geometry at various depth levels. This representation is inspired by classical semantic segmentation, but extends segmentation planes to various depth levels. OPlanes can be used to generate an occupancy grid, from which the mesh  $\mathcal{M}$ is obtained via a marching cube~\cite{supplorensen1987marching} algorithm. We illustrate the framework in \figref{fig: example img}.
    
In the following we first introduce OPlanes in~\secref{sec: oplane}.
Subsequently, \secref{sec: oplane pred} details the developed deep net to predict OPlanes, while  training of the deep net is discussed in~\secref{sec: oplane train}.
Finally, \secref{sec: oplane to mesh} provides details about generation of the mesh from the predicted OPlanes.

\subsection{Occupancy Planes (OPlanes) Representation}\label{sec: oplane}

Given an image capturing a human of interest, \emph{occupancy planes} (OPlanes) store the occupancy information of that human in the camera's view frustum.
For this, the OPlanes representation consists of several 2D images, each of which store the mesh occupancy at a specific fronto-parallel slice through the camera's view frustum.

Concretely, let $[z_\text{min}, z_\text{max}]$ be the range of depth we are interested in,~\ie, $z_\text{min}, z_\text{max}$ are the near-plane and far-plane of the view frustum of interest. Further, let the set $\mathcal{Z}_N \triangleq \{ z_1, \dots, z_N \,\vert\, z_\text{min} \leq z_i \leq z_\text{max}, \forall i \}$ contain the sampled depths of interest.
The OPlanes representation $\mathcal{O}_{\mathcal{Z}_N}^{H\times W}$ for the depths of interest stored in $\mathcal{Z}_N$ refers to the set of planes
\begin{align}
    \mathcal{O}_{\mathcal{Z}_N}^{H\times W} \triangleq \left\{ O_{z_1}^{H\times W}, O_{z_2}^{H\times W}, \dots, O_{z_N}^{H\times W} \,\vert\, z_i \in \mathcal{Z}_N  \right\}.
\end{align}
Each OPlane $O_{z}^{H\times W} \in \{0, 1\}^{H \times W}$ is a binary image of height $H$ and width $W$.

To compute the ground-truth binary values of the occupancy plane $O_{z}^{H\times W}$ at depth $z$, let $[x, y, 1]$ be a homogeneous pixel coordinate on the given image $I$. Given a depth $z$ of interest, the homogeneous pixel coordinate can be unprojected into the 3D space coordinate $[x_z, y_z, z] = z \cdot \pi^{-1}([x, y, 1])$, where $\pi(\cdot)$ denotes the perspective projection.
Note, this unprojection differs from prior human mesh reconstruction works~\cite{Zheng2019DeepHuman3H, Saito2019PIFuPI, Saito2020PIFuHDMP, He2020GeoPIFuGA} that assume an orthogonal projection with a weak-perspective camera. Instead, we utilize a perspective camera for more general use cases.
    
From  3D meshes available in the training data we obtain a 3D point's ground-truth occupancy value as follows:
    \begin{align}
        o([x_z, y_z, z]) =
       \begin{cases}
       1, \;\text{if $[x_z, y_z, z]$ inside the object},\\
       0, \;\text{otherwise}.
       \end{cases}
    \end{align}
The value of the occupancy plane $O_{z}^{H\times W} \in \{0, 1\}^{H \times W}$ at depth $z$ and at pixel location $x, y$ can be obtained from the ground-truth occupancy value via
    \begin{align}
        O_{z}^{H\times W}[x, y] = o([x_z, y_z, z]),
    \end{align}
    where $O_{z}^{H\times W}[x, y]$ denotes the occupancy plane value of pixel $[x, y]$.

\subsection{Occupancy Planes Prediction}\label{sec: oplane pred}

At test time, ground-truth meshes are not available. Instead we are interested in predicting the OPlanes from 1) a given RGB image $I_\text{RGB} \in \mathbb{R}^{H \times W \times 3}$ illustrating a human, 2) a depth map $\texttt{Depth} \in \mathbb{R}^{H \times W}$,  3) a mask $\texttt{Mask} \in \{0, 1\}^{H \times W}$, and 4) the calibrated camera's perspective projection $\pi$.
Specifically, let $H_O\times W_O$ be the operating resolution where $H_O \leq H$ and $W_O \leq W$.
We use a deep net to predict $N$ occupancy planes $\widehat{\mathcal{O}}^{H_O\times W_O}= \{ \widehat{O}_{z_i}^{H_O\times W_O} \}_{i=1}^N$ at various depth levels $z_i$ $\forall i\in\{1, \dots, N\}$, via
\begin{align}
    \widehat{O}_{z_i}^{H_O\times W_O} = f_\text{spatial}([\mathcal{F}_\text{RGB}^{H_O\times W_O}; \mathcal{F}_{z_i}^{H_O\times W_O}] ). \label{eq: O_zi final}
\end{align}
Here, $[\cdot; \cdot]$ denotes the concatenation operation along the channel dimension.

In order to resolve the depth ambiguity, we design $f_\text{spatial} (\cdot)$ to be a simple fully convolutional network that aims to fuse spatial neighborhood information within each occupancy plane prediction $\widehat{O}_{z_i}^{H_O\times W_O}$. Note that this design differs from prior work, which predicts the occupancy for each point independently. In contrast, we find  spatial neighborhood information is useful to improve occupancy prediction accuracy.

For an accurate prediction, the fully convolutional net $f_\text{spatial} (\cdot)$ operates on image features $\mathcal{F}_\text{RGB}^{H_O\times W_O} \in \mathbb{R}^{H_O \times W_O \times C}$ and depth features $\mathcal{F}_{z_i}^{H_O\times W_O}  \in \mathbb{R}^{H_O \times W_O \times C}$. In the following we discuss the deep nets to compute the image features $\mathcal{F}_\text{RGB}^{H_O\times W_O}$ and the depth features $\mathcal{F}_{z_i}^{H_O\times W_O}$.
    
\noindent\textbf{Image feature $\mathcal{F}_\text{RGB}$.}
The image feature $\mathcal{F}_\text{RGB}^{H_O\times W_O}$ is obtained by bilinearly upsampling a low-resolution feature map to the operating resolution $H_O\times W_O$. Concretely, 
        \begin{align}
            \mathcal{F}_\text{RGB}^{H_O\times W_O} = \texttt{UpSample}_{h_O\times w_O \rightarrow H_O\times W_O} (\mathcal{F}_\text{RGB}^{h_O \times w_O}),
        \end{align}
         where $\mathcal{F}_\text{RGB}^{h_O \times w_O} \in \mathbb{R}^{h_O \times w_O \times C}$ is the RGB feature at the coarse resolution of $h_O \times w_O$.
        $\texttt{UpSample}_{h_O\times w_O \rightarrow H_O\times W_O}$ refers to the standard bilinear upsampling.
        
The coarse resolution RGB feature  is obtained via 
        \begin{align}
            \mathcal{F}_\text{RGB}^{h_O \times w_O} = f_\text{RGB} (f_\text{FPN} (\hat{I}_\text{RGB})), \label{eq: rgb feat}
        \end{align}
where $\hat{I}_\text{RGB} \in \mathbb{R}^{H\times W \times5}$ is the concatenation of $I_\text{RGB}$ and two simple features (see appendix). $f_\text{FPN}$ is the Feature Pyramid Network (FPN) backbone~\cite{Lin2017FeaturePN} and $f_\text{RGB}$ is another fully-convolutional network for further processing.
    
\noindent\textbf{Depth feature $\mathcal{F}_{z_i}$.}
The depth feature $\mathcal{F}_{z_i}^{H_O\times W_O}$ for an occupancy plane at depth $z_i$ encodes for every pixel $[x,y]$ the difference between the query depth $z_i$ and the depth at which the object first intersects with the camera ray. 
Concretely, we obtain the depth feature via
        \begin{align}
            \mathcal{F}_{z_i}^{H_O\times W_O} = f_\text{depth} (I_{z_i}^{H_O\times W_O}), \label{eq: depth feat}
        \end{align}
        where $f_\text{depth}$ is a fully convolutional network to process the depth difference image $I_{z_i}^{H_O\times W_O}$.
        
The depth difference image $I_{z_i}^{H_O\times W_O}$ is constructed to capture the  difference between the query depth $z_i$ and the depth at which the object first intersects with the camera ray.~\Ie, for each pixel $[x, y]$,
\begin{align}
    I_{z_i}^{H_O\times W_O} [x, y] = \texttt{PE} (z_i - \texttt{Depth}[x, y]),\label{eq: z diff img}
\end{align}
where $\texttt{PE}(\cdot)$ is the positional encoding operation~\cite{Vaswani2017AttentionIA}.
Intuitively, the depth difference image $I_{z_i}$ represents how far every point on the  plane at depth $z_i$ is behind or in front of the front surface of the observed human.

\subsection{Training}\label{sec: oplane train}

The developed deep net to predict OPlanes is fully differentiable. We use $\theta$ to subsume all trainable parameters within the spatial network $f_\text{spatial}$ (\equref{eq: O_zi final}), the FPN network $f_\text{FPN}$,  the RGB network $f_\text{RGB}$ (\equref{eq: rgb feat}), and the depth network $f_\text{depth}$ (\equref{eq: depth feat}). Further, we use $\widehat{\mathcal{O}}_\theta$ to refer to the predicted occupancy planes when using the parameter vector $\theta$.
We train the deep net to predict OPlanes end-to-end with two losses by addressing
\begin{align}
    \min_\theta \mathcal{L}_\theta^{H_O \times W_O} + \mathcal{L}_\theta^{h_O \times w_O}.\label{eq: loss}
\end{align}
Here, $\mathcal{L}_\theta^{H_O \times W_O}$ is the loss computed at the final prediction  resolution of $H_O \times W_O$, while $\mathcal{L}_\theta^{h_O \times w_O}$ is used to supervise intermediate features at the resolution of $h_O \times w_O$. We discuss both losses next.
    
\noindent\textbf{Final prediction supervision via $\mathcal{L}_\theta^{H_O \times W_O}$.}
During training, we randomly sample $N$ depth values from the view frustum range $[z_\text{min}, z_\text{max}]$ to obtain the set of depth values of interest $\mathcal{Z}_N$ (\secref{sec: oplane}).
For this, we use $z_\text{min} = \min \{ \texttt{Depth}[x, y] \,\vert\, \texttt{Mask}[x, y] == 1 \}$ by only considering depth information within the target mask. Essentially, we find the depth value that is closest to the camera.
We set $z_\text{max} = z_\text{min} + z_\text{range}$, where $z_\text{range}$ marks the depth range we are interested in. During training, $z_\text{range}$ is computed from the ground-truth range which covers the target mesh. During inference, we set $z_\text{range} = 2 $ meters to cover the shapes and gestures of most humans.

The high resolution supervision loss $\mathcal{L}_\theta^{H_O \times W_O}$ consists of two terms. Namely $\mathcal{L}_\theta^{H_O \times W_O} \triangleq$
\begin{align}
    &\lambda_\text{BCE} \cdot \mathcal{L}_\text{BCE}(\mathcal{O}^{H_O \times W_O}, \widehat{\mathcal{O}}_\theta^{H_O \times W_O}, \texttt{Mask}^{H_O \times W_O}, z_\text{min}, z_\text{max}) + \nonumber \\
    &\lambda_\text{DICE} \cdot \mathcal{L}_\text{DICE}(\mathcal{O}^{H_O \times W_O}, \widehat{\mathcal{O}}_\theta^{H_O \times W_O}, \texttt{Mask}^{H_O \times W_O}, z_\text{min}, z_\text{max}).\label{eq: fine loss}
\end{align}
Here, $\mathcal{L}_\text{BCE}$ is the binary cross entropy (BCE) loss while $\mathcal{L}_\text{DICE}$ is the DICE loss~\cite{Milletari2016VNetFC}. Both losses operate on the ground-truth OPlanes ${\cal O}^{H_O \times W_O}$ downsampled from the original resolution $H\times W$, the OPlanes $\widehat{\cal O}_\theta^{H_O \times W_O}$ predicted with the current deep net parameters $\theta$, and the human mask $\texttt{Mask}^{H_O \times W_O}$ downsampled from the raw mask.
Note, we only consider points behind the human's front surface when computing the loss, \ie,  on a plane $\widehat{O}_{z_i}$, we only consider $\{[x, y] \vert z_i \geq \texttt{Detph}[x, y] \}$. 
For readability, we drop the superscript ${H_O \times W_O}$ in the following.
The BCE loss is computed via $\mathcal{L}_\text{BCE} =$
\begin{align}
    &\frac{1}{\vert \mathcal{Z}_N\vert \!\cdot\! \texttt{Sum}(\texttt{Mask}) }\hspace{-0.4cm}
    \sum\limits_{\substack{z_i \in \mathcal{Z}_N \\ x,y:\texttt{Mask}[x, y] = 1 }} \biggl(\! O_{z_i}[x, y] \cdot \log \widehat{O}_{z_i}[x, y]  \nonumber \\
    &\quad\quad\quad + (1 - O_{z_i}[x, y]) \cdot \log (1 - \widehat{O}_{z_i}[x, y]) \!\biggr),
\end{align}
where $\texttt{Sum}(\texttt{Mask})$ is the number of pixels within the target's segmentation mask and
$x,y:\texttt{Mask}[x, y] = 1$ emphasizes that we only compute BCE loss on  pixels within the mask.
    
Moreover, thanks to the occupancy plane representation inspired by semantic segmentation tasks, we can utilize the DICE loss from the semantic segmentation community to supervise the occupancy training. 
Specifically, we use $\mathcal{L}_\text{DICE} =$
        \begin{align}
             \frac{1}{\vert \mathcal{Z}_N\vert} \sum\limits_{\substack{z_i \in \mathcal{Z}_N}} \frac{2 \cdot \texttt{Sum}( \texttt{Mask} \cdot O_{z_i} \cdot \widehat{O}_{z_i})}{\texttt{Sum}(\texttt{Mask} \cdot O_{z_i}) + \texttt{Sum}(\texttt{Mask} \cdot \widehat{O}_{z_i})}.
        \end{align}
This is useful because there can be a strong imbalance between the number of positive and negative labels in an OPlane  $O_{z_i}$ due to  human gestures. The DICE loss has been shown to compellingly deal with such  situations~\cite{Milletari2016VNetFC}. 
    
\noindent\textbf{Intermediate feature supervision via $\mathcal{L}_\theta^{h_O \times w_O}$.}
Besides supervision of the final occupancy image $\widehat{O}_{z_i}$ discussed in the preceding section, we also supervise the intermediate features $\mathcal{F}_\text{RGB}^{h_O\times w_O}$ (\equref{eq: rgb feat}) via the loss $\mathcal{L}_\theta^{h_O \times w_O}$. 
Analogously to the high-resolution loss, we use two terms, \ie, $\mathcal{L}_\theta^{h_O \times w_O} \triangleq$
        \begin{align}
            &\lambda_\text{BCE} \cdot \mathcal{L}_\text{BCE}(\mathcal{O}^{h_O \times w_O}, \widehat{\mathcal{O}}_\theta^{h_O \times w_O}, \texttt{Mask}^{h_O \times w_O}, z_\text{min}, z_\text{max}) + \nonumber \\
            &\lambda_\text{DICE} \cdot \mathcal{L}_\text{DICE}(\mathcal{O}^{h_O \times w_O}, \widehat{\mathcal{O}}_\theta^{h_O \times w_O}, \texttt{Mask}^{h_O \times w_O}, z_\text{min}, z_\text{max}).\label{eq: coarse loss}
        \end{align}
Different from the high-resolution representation, we predict the OPlanes representation at the coarse resolution $h_O \times w_O$ via
        \begin{align}
            \widehat{O}_{z_i}^{h_O \times w_O}[x, y] = \langle \mathcal{F}_\text{RGB}^{h_O \times w_O}[x, y, \cdot],\, \mathcal{F}_{z_i}^{h_O \times w_O}[x, y, \cdot] \rangle,
        \end{align}
        where $\langle \cdot, \cdot \rangle$ is the inner-product operation and $\mathcal{F}_\text{RGB}^{h_O \times w_O}[x, y, \cdot]$ represents the feature vector at the pixel location $[x, y]$. 
To obtain $\mathcal{F}_{z_i}^{h_O \times w_O}$, we feed the downsampled difference image $I_{z_i}^{h_O \times w_O}$ into $f_\text{depth}$.
Intuitively, we use the inner product to  encourage the image feature $\mathcal{F}_\text{RGB}^{h_O \times w_O}$ to be strongly correlated to information from the depth feature $\mathcal{F}_{z_i}^{h_O \times w_O}$.

\subsection{Inference}\label{sec: oplane to mesh}

During inference, to reconstruct a mesh from predicted OPlanes $\widehat{\cal O}$, we first establish an occupancy grid before running a marching cube~\cite{supplorensen1987marching} algorithm to extract the isosurface.
Specifically, we uniformly sample $N$ depths in the view frustum between depth range $[z_\text{min}, z_\text{min} + 2.0]$,~\eg, $N = 256$. Here 2.0 is a heuristic depth range which covers most human poses (\secref{sec: oplane train}).
The network predicts an occupancy for each pixel on those $N$ planes. Importantly, since  OPlanes represent occupancy corresponding to slices through the view frustum, a marching cube algorithm is not directly applicable. Instead, we first establish a voxel grid  to cover the view frustum between $[z_\text{min}, z_\text{min} + 2.0]$. Each voxel's occupancy is sampled from the predicted OPlanes before a marching cube method is used.
We  emphasize that the number of planes do not need to be the same during training and inference, which we will show later. This  ensures that the OPlanes representation is memory efficient at training time while enabling accurate reconstruction at inference time. 

\section{Experiments}

\subsection{Implementation Details}\label{sec: implement}

Here we introduce key implementation details. Please see the appendix for more information.
During training, the input has a resolution of $H = 512$ and $W = 512$. We operate at $H_O = 256$, $W_O = 256$, while the intermediate resolution is $h_O = 128$ and $w_O = 128$. During training, for each mesh,  we randomly sample $N = 10$ planes in the range of $[z_\text{min}, z_\text{max}]$ at each training iteration.~\Ie, the set $\mathcal{Z}_N$ contains 10 depth values.
As mentioned in~\secref{sec: oplane train}, during training, we set $z_\text{max}$ to be the ground-truth mesh's furthest depth.

The four deep nets, which we detail next, are mostly convolutional. We use \textit{(in, out, k)} to denote the input/output channels and the kernel size %
of a convolutional layer.

\noindent\textbf{Spatial network $f_\text{spatial}$} (\equref{eq: O_zi final}): It's a three-layer convolutional neural net (CNN) with a configuration of (256, 128, 3), (128, 128, 3), (128, 1, 1). We use group norm~\cite{Wu2018GroupN} and ReLU activation.

\noindent\textbf{Feature pyramid network $f_\text{FPN}$} (\equref{eq: rgb feat}): We use ResNet50~\cite{He2016DeepRL} as the backbone of our FPN network. We use the output of each stage's last residual block as introduced in~\cite{Lin2017FeaturePN}. The final output of this FPN has 256 channels and a resolution of $\frac{H}{4} \times \frac{W}{4}$.

\noindent\textbf{RGB network $f_\text{RGB}$} (\equref{eq: rgb feat}): It's a three-layer CNN with a configuration of (256, 128, 3), (128, 128, 3), (128, 128, 1). We use group norm~\cite{Wu2018GroupN} and ReLU activation.

\noindent\textbf{Positional encoding \texttt{PE}} (\equref{eq: z diff img}): We follow~\cite{Vaswani2017AttentionIA} to define $\texttt{PE}(\text{pos}) =$
\begin{align}
     \left( \texttt{PE}_0(\text{pos}), \texttt{PE}_1(\text{pos}), \dots, \texttt{PE}_{63}(\text{pos}), \texttt{PE}_{64}(\text{pos}) \right),
\end{align}
where $\texttt{PE}_{2t}(\text{pos}) = \sin(\frac{50 \cdot \text{pos}}{200^{2t / 64}})$ and $\texttt{PE}_{2t + 1}(\text{pos}) = \cos(\frac{50 \cdot \text{pos}}{200^{2t / 64}})$.

\noindent\textbf{Depth difference network $f_\text{depth}$} (\equref{eq: depth feat}): It's a two-layer CNN with a configuration of (64, 128, 1), (128, 128, 1). We use group norm~\cite{Wu2018GroupN} and ReLU activation.

To train the networks, we use the Adam~\cite{Kingma2015AdamAM} optimizer with a learning rate of 0.001. We set $\lambda_\text{BCE} = 1.0$ and $\lambda_\text{DICE} = 1.0$ (\equref{eq: fine loss} and~\equref{eq: coarse loss}). We set the batch size to 4 and train for 15 epochs. It takes around 22 hours to complete the training using an AMD EPYC 7543 32-Core Processor and an Nvidia RTX A6000 GPU.

\subsection{Experimental Setup}

\begin{table*}[t]
\renewcommand{\arraystretch}{1.0}
\begin{adjustwidth}{0.0cm}{}
\renewcommand\theadfont{}
\centering
\captionsetup{width=\linewidth}
\caption{\textbf{Quantitative results}.
Each result averages three runs with different seeds and is reported in the format of $\texttt{mean}{\scriptsize \pm \texttt{std}}$.
OPlanes improve upon PIFuHD by a  margin ($1^\text{st}$~\vs~$6^\text{th}$ row) and outperform IF-Net in almost all metrics ($2^\text{nd}$~\vs~$6^\text{th}$ row).
We also verify the design choices via an ablation study reported in the $3^\text{rd}$ to $5^\text{th}$ row.
For all OPlanes results, we predict occupancy using 256 planes per mesh during inference, while  using  10 or less planes per mesh when training.
}
\label{tab: quant}
\vspace{-0.2cm}
\begin{adjustbox}{width=0.8\textwidth,center}
\setlength{\tabcolsep}{3pt}
{
\begin{tabular}{lcccccrrr} 
\toprule
 & & {\makecell{OPlane}} & \makecell{$f_\text{spatial}$\\Kernel Size} & \makecell{$\mathcal{L}_\theta^{h_O \times w_O}$} &  \makecell{\#Planes\\in Train} & IoU$\uparrow$  & Cham-$\mathcal{L}_1 \downarrow$ & \makecell{Normal\\Consist.} $\uparrow$ \\
\midrule
1 & PIFuHD~\cite{Saito2020PIFuHDMP} & \xmark & - & - & - & 0.428 & 0.332 & 0.677 \\
2 & IF-Net~\cite{Chibane2020ImplicitFI} & \xmark & - & - & - & 0.584  & 0.216  & \textbf{0.802} \\
\midrule
3 & NoNeighborInfo & \cmark  & $1\times1$ & \gcmark & \textcolor{light-gray}{5} & 0.679{\scriptsize$\pm$0.013}  & 0.158{\scriptsize$\pm$0.007}  & 0.738{\scriptsize$\pm$0.005} \\
4 & NoInterSupervision & \cmark  & \textcolor{light-gray}{$3\times3$} & \xmark  & \textcolor{light-gray}{5}  & 0.681{\scriptsize$\pm$0.013}  & 0.161{\scriptsize$\pm$0.008}  & 0.739{\scriptsize$\pm$0.005}   \\
5 & LessPlanes & \cmark & \textcolor{light-gray}{$3\times3$} & \gcmark & 5 & 0.684{\scriptsize$\pm$0.013}  & 0.158{\scriptsize$\pm$0.008}  & 0.747{\scriptsize$\pm$0.005}   \\
\midrule
\belowrulesepcolor{highlightRowColor} 
\rowcolor{highlightRowColor}
6 & OursFull & \cmark  & $3\times3$ & \cmark & 10 & \textbf{0.691{\scriptsize$\pm$0.013}}  & \textbf{0.155{\scriptsize$\pm$0.008}}  & 0.749{\scriptsize$\pm$0.005} \\
\toprule
\end{tabular}
}
\end{adjustbox}
\end{adjustwidth}
\vspace{-0.3cm}
\end{table*}

\textbf{Dataset.} We utilize S3D~\cite{Hu2021SAILVOS3A} to train our OPlanes-based human reconstruction model. S3D is a photo-realistic synthetic dataset built on the game GTA-V, providing ground-truth meshes together with masks and depths. To construct our train and test set, we sample 27588 and 4300 meshes from its train and validation split respectively.
This dataset differs from counterparts in prior works~\cite{Saito2019PIFuPI, Saito2020PIFuHDMP, He2020GeoPIFuGA, Alldieck2022PhotorealisticM3}:
there are no constraints on the appearance of humans in the images. In our dataset, humans  appear with any gestures, any sizes, any position, and any level of occlusion. In contrast, humans in datasets of prior work  usually appear in an upright position and are mostly centered in an image while exhibiting little to no occlusion.
We think this setup  strengthens the generalization ability. See~\figref{fig: s3d qualitative} for some examples.

\noindent\textbf{Baselines.}
We compare to PIFuHD~\cite{Saito2020PIFuHDMP} and IF-Net~\cite{Chibane2020ImplicitFI}. \textbf{1) PIFuHD:} since there is no training code available, we test with the officially-released checkpoints. Following the author's suggestion in the public code repository\footnote{\url{https://github.com/facebookresearch/pifuhd}} to improve the reconstruction quality, we 1.1) remove the background with the ground-truth mask; 1.2) apply human pose detection~\cite{osokin2018lightweight_openpose} and crop the image accordingly to place the human of interest in the center of the image.
\textbf{2) IF-Net: } we evaluate with the officially-released checkpoint.
IF-Net uses a 3D voxel grid representation. 
We set the resolution of the grid to  256 to align with the pretrained checkpoint.

\noindent\textbf{Evaluation metrics.} We focus on evaluating the quality of the reconstructed geometry.  Following prior works~\cite{Mescheder2019OccupancyNL,Saito2019PIFuPI, Saito2020PIFuHDMP, He2020GeoPIFuGA, Alldieck2022PhotorealisticM3, Huang2020ARCHAR, He2021ARCHAC}, we report the Volumetric Intersection over Union (IoU), the bi-directional Chamfer-$\mathcal{L}_1$ distance, and the Normal Consistency. Please refer to the supplementary material of~\cite{Mescheder2019OccupancyNL} for more details on these metrics. 
To compute the IoU, we need a finite space to sample points.
Since humans in our data appear anywhere in 3D space, the implicit assumption of prior works~\cite{Saito2019PIFuPI, Saito2020PIFuHDMP, He2020GeoPIFuGA} that there exists a fixed bounding box for all objects does not hold.
Instead, we use the view frustum between depth $z_\text{min}$ and $z_\text{max}$ as the bounding box. Note, for evaluation purposes, $z_\text{max}$ utilizes the heuristic $z_\text{range}$ of  2.0 meters (\secref{sec: oplane train}). We sample 100k points for an unbiased estimation.
When computing the Chamfer distance, we need to avoid that the final aggregated results are skewed by a scale discrepancy between different objects.
We follow~\cite{Fan2017APS, Mescheder2019OccupancyNL} and let $\frac{1}{10}$ of each object's ground-truth bounding box's longest edge  correspond to a unit of one when computing Chamfer-$\mathcal{L}_1$. %
To resolve the discrepancy between the orthogonal projection and the perspective projection, we utilize the iterative-closest-point (ICP)~\cite{Besl1992AMF} algorithm to register  the reconstruction of baselines to the ground-truth, following~\cite{Alldieck2022PhotorealisticM3}. ICP is not applied to our OPlanes method since we directly reconstruct the human in the camera coordinate system.

\subsection{Quantitative Results}

In~\tabref{tab: quant} we provide quantitative results, comparing to baselines in the $1^\text{st}$/$2^\text{nd}$~\vs~$6^\text{th}$ row. For a fair comparison when computing the results, we reconstruct the final geometry in a $256^3$ grid. Although 256 OPlanes are inferred, we train with only 10 planes per mesh in each iteration.

\noindent\textbf{PIFuHD~\cite{Saito2020PIFuHDMP}:} The OPlanes representation outperforms the PIFuHD results by a margin. Specifically, our results exhibit a larger volume overlap with the ground-truth (0.691~\vs~0.428 on IoU, $\uparrow$ is better), more completeness and accuracy (0.155~\vs~0.332 on Chamfer distance, $\downarrow$ is better), and more fine-grained details (0.749~\vs~0.677 on normal consistency, $\uparrow$ is better).

\begin{figure*}[!t]
    \centering
    \captionsetup[subfigure]{width=\textwidth}
        \centering
        \includegraphics[width=0.8\textwidth]{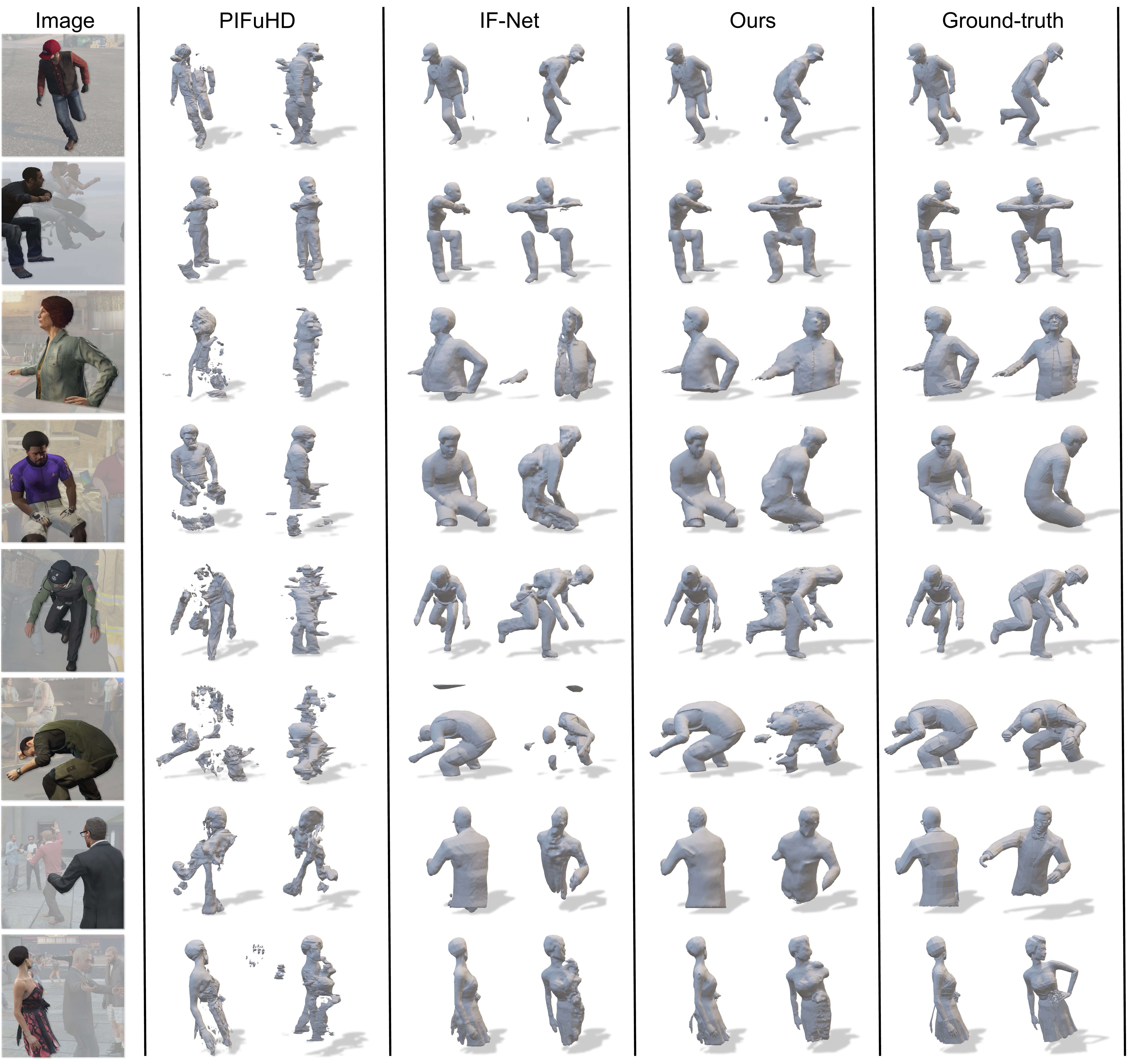}
        \captionsetup{width=\textwidth}
        \vspace{-0.2cm}
        \caption{Qualitative results on S3D~\cite{Hu2021SAILVOS3A}. For each reconstruction, we show two views. PIFuHD~\cite{Saito2020PIFuHDMP} and IF-Net~\cite{Chibane2020ImplicitFI} struggle to obtain consistent geometry if parts of the human are invisible, while an OPlanes model faithfully reconstructs the visible portion.}
        \label{fig: s3d qualitative}
\vspace{-0.4cm}
\end{figure*}

\begin{figure}[t]
    \centering
    \captionsetup[subfigure]{width=\columnwidth}
        \centering
        \hspace*{0.1cm}\includegraphics[width=\columnwidth]{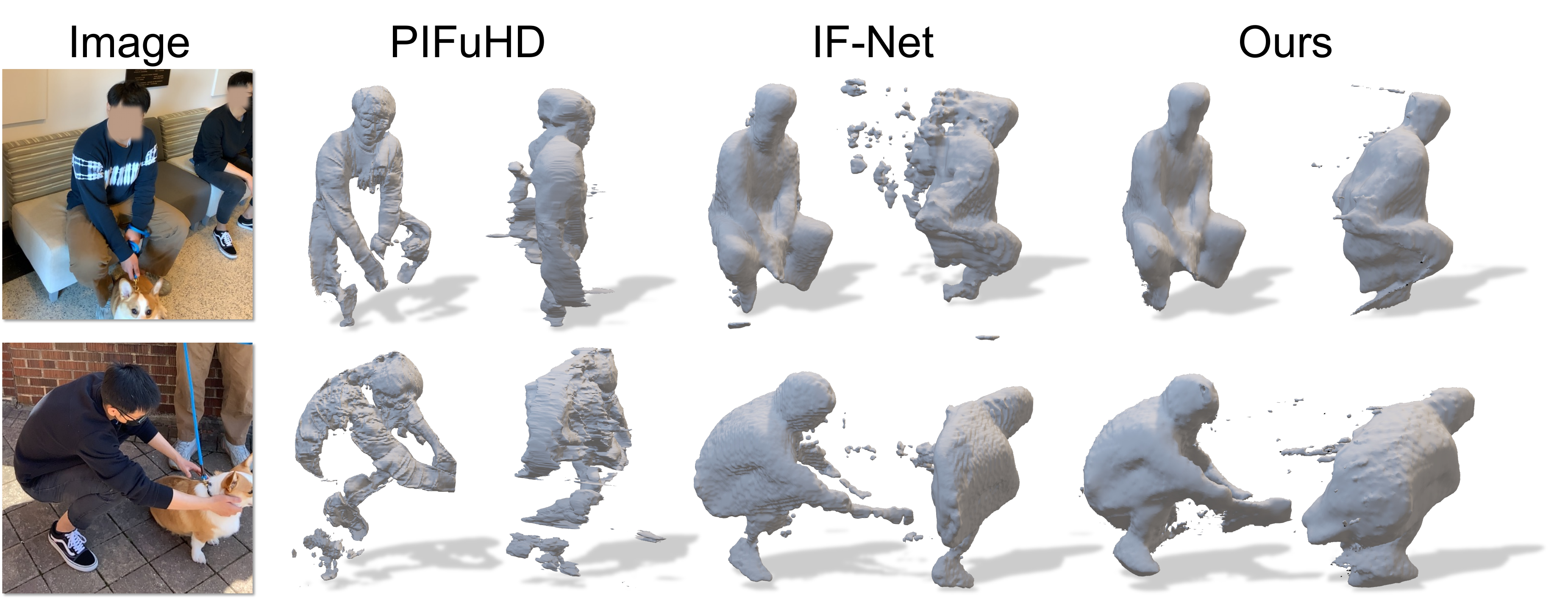}
        \captionsetup{width=\columnwidth}
         \vspace{-0.5cm}
        \caption{Transfer results on real-world RGB-D data captured with a 2020 iPad Pro.}
        \label{fig: transfer apple}
\vspace{-0.5cm}
\end{figure}

\noindent\textbf{IF-Net~\cite{Chibane2020ImplicitFI}:}  We also compare to the depth-based single-view reconstruction approach IF-Net.
The results are presented in row 2 \vs 6 in~\tabref{tab: quant}.
We find that IF-Net struggles to reconstruct humans which are partly occluded or outside the field-of-view (see~\figref{fig: s3d qualitative} and~\figref{fig: transfer apple} for some examples).
More importantly, we observe IF-Net to yield inferior results with respect to IoU (0.584~\vs~0.691, $\uparrow$ is better) and Chamfer distance (0.216~\vs~0.155, $\downarrow$ is better).
Notably, we find the high normal consistency of IF-Net to be due to the high-resolution voxel grid, which provides more details.

\subsection{Analysis}

To verify design choices, we conduct ablation studies. We report the results in \tabref{tab: quant}'s $3^\text{rd}$ to $5^\text{th}$ row.

\noindent\textbf{Per-point classification is not all you need:} To understand whether neighboring information is needed, we replace the $3\times 3$ kernel in $f_\text{spatial}$ (\equref{eq: O_zi final}, \secref{sec: implement}) with a $1\times 1$ kernel, which essentially conducts per-point classification for each pixel on the OPlane. Comparing the $3^\text{rd}$~\vs~$5^\text{th}$ row in~\tabref{tab: quant} corroborates the importance of context  as per-point classification yields inferior results. This shows that the conventional way to treat  shape reconstruction as a point classification problem~\cite{Saito2019PIFuPI, Saito2020PIFuHDMP, He2020GeoPIFuGA} may be suboptimal. Specifically, without directly taking into account the context information, we observe lower IoU (0.679~\vs~0.684) and less normal consistency (0.738~\vs~0.747). 

\noindent\textbf{Intermediate supervision is important:} To understand whether the supervision of intermediate features is needed, we train our OPlanes without $\mathcal{L}_\theta^{h_O \times w_O}$ (\equref{eq: coarse loss}). The results in \tabref{tab: quant}'s $4^\text{th}$~\vs~$5^\text{th}$ row verify the benefits of intermediate supervision. Concretely, with intermediate feature supervision, we obtain a better IoU (0.684~\vs~0.681, $\uparrow$ is better), an improved Chamfer distance (0.158~\vs~0.161, $\downarrow$ is better), and a better normal consistency (0.747~\vs~0.739, $\uparrow$ is better).

\noindent\textbf{Training with more planes is beneficial:} We are curious about whether training with less planes harms the performance of our OPlanes model. For this, we sample only 5 planes per mesh when training the OPlanes model. The results in the  $5^\text{th}$~\vs~$6^\text{th}$ row in~\tabref{tab: quant} demonstrate that training with more planes yields better results. Concretely, with more planes, we obtain better IoU (0.691~\vs~0.684, $\uparrow$ is better), smaller Chamfer distance (0.155~\vs~0.158, $\downarrow$ is better), and better normal consistency (0.749~\vs~0.747, $\uparrow$ is better).

\subsection{Qualitative Results}

\textbf{S3D.}
We provide qualitative results in~\figref{fig: s3d qualitative}. OPlanes successfully handle various human gestures and different levels of visibility while  PIFuHD and IF-Net fail in those situations.

\noindent\textbf{Transferring Results to Real-World RGB-D Data.}
In \figref{fig: transfer apple} we use real-world data collected in the wild to compare to PIFuHD and IF-Net.  OPlanes results are obtained by directly applying the proposed OPlanes model trained on S3D, without fine-tuning or other adjustments. 
For this result, we use a 2020 iPad Pro equipped with a LiDAR sensor~\cite{arkit2021} and develop an iOS app to acquire the RGB-D images and camera matrices.
The human masks are obtained by feeding RGB images into a Mask2Former~\cite{Cheng2021MaskedattentionMT}.
1) We observe  PIFuHD results to be noisy and to contain holes. 
2)  For humans that are only partially visible, IF-Net seems to struggle.
Our model benefits from the OPlanes representation which  better exploits  correlations within a plane. For this reason OPlanes better capture the human shape despite the model being trained on  synthetic data.

\section{Conclusion}\label{sec:conc}
We propose and study the occupancy planes (OPlanes) representation for reconstruction of 3D shapes of humans from a single RGB-D image. The resolution of OPlanes is more flexible than that of a classical voxel grid due to the implicit prediction of an entire plane. Moreover, prediction of an entire plane enables the model to benefit from correlations between predictions, which is harder to achieve for models which use implicit functions for individual 3D points. Due to these benefits we find OPlanes to excel in challenging situations, particularly for occluded or partially visible humans.

\noindent\textbf{Acknowledgements.} Work supported in part by NSF under Grants 1718221, 2008387, 2045586, 2106825, MRI 1725729, and NIFA award 2020-67021-32799. Thanks to NVIDIA for providing a GPU for debugging.

\bibliography{ref}

\clearpage
\beginsupplement
\appendix

\onecolumn

\section*{\Large\centering Supplementary Material: \\Occupancy Planes for Single-view RGB-D Human Reconstruction}\vspace{0.2cm}

This supplementary material is structured as follows:
\begin{enumerate}
    \item \secref{sec: supp implement}: Implementation details
    \item \secref{sec: supp quantitative}: Additional quantitative results
\end{enumerate}

\section{Implementation Details}\label{sec: supp implement}

\subsection{Input to Image Feature Extractor}

To extract the image features $\mathcal{F}_\text{RGB}^{h_O \times w_O} = f_\text{RGB} (f_\text{FPN} (\hat{I}_\text{RGB}))$ (see \equref{eq: rgb feat}), instead of feeding the raw RGB image $I_\text{RGB} \in \mathbb{R}^{H\times W\times 3}$ into the FPN backbone, we first concatenate the image $I_\text{RGB}$ with two simply-processed one-channel features which are concatenated to the image along the channel dimension. We therefore use $\hat{I}_\text{RGB} \in \mathbb{R}^{H\times W\times 5}$, which will be fed into the FPN,~\ie,~$\mathcal{F}_\text{RGB}^{h_O \times w_O} = f_\text{RGB} (f_\text{FPN} (\hat{I}_\text{RGB}))$.
The two one-channel features are: 1) for each pixel, we compute the distance to the visibility mask's boundary; 2) we detect edges with the help of a Farid filter~\cite{Farid2004DifferentiationOD}.

\subsection{Visualization}

Even though occupancy planes learn a high-quality inductive bias for single-view RGB-D human reconstruction,  floating artifacts behind the visible surfaces are possible as inferring invisible parts is an ill-posed problem. For a smooth visualization, we utilize 1) the smoothing function in \texttt{PyMCubes}~\footnote{\url{https://github.com/pmneila/PyMCubes}}; 2) the GraphCut algorithm~\cite{Boykov2004AnEC} in \texttt{medpy}~\footnote{\url{https://github.com/loli/medpy/}}. Importantly, note that we only apply this post-processing for visualization purposes and we \emph{never} use this post-processing when reporting quantitative results.

\section{More Quantitative Results}\label{sec: supp quantitative}

\subsection{Performance across Various Visibility Levels}

Since  OPlanes can deal with humans of various visibilities, we are interested in understanding how the proposed approach performs across different partial visibility levels. We present results with respect to different visibility levels in~\tabref{tab: supp vis level}.
To compute the visibility we use three steps: 1) we uniformly sample 100k points within the  complete mesh of the human; 2) we project those 100k 3D points onto the 2D image and count the number of points which are in view; 3) the level of partial visibility is computed as the ratio of in-view points,~\ie,~the number of in-view points divided by 100k.
We also explicitly consider the fully visible humans in the $4^\text{th}$ row of~\tabref{tab: supp vis level}.
Results for different visibility ranges are provided in the $1^\text{st}$ to $3^\text{rd}$ row of \tabref{tab: supp vis level}.
As expected, the more visible the human, the better the model performs.
Specifically, comparing full visibility to low visibility ($4^\text{th}$~\vs~$1^\text{st}$ row), we obtain higher IoU (0.707~\vs~0.668), smaller Chamfer distance (0.109~\vs~0.289), and more normal consistency (0.759~\vs~0.703). However, it is notable that the drop in performance is not very severe.

To verify this, we also report IF-Net and PIFuHD results for each visibility range in~\tabref{tab: supp vis level}. Specifically, comparing the $4^\text{th}$~\vs~$1^\text{st}$ row, we observe: 
1) for IoU ($\uparrow$ is better), IF-Net's performance drops from 0.644 to 0.365 and PIFuHD results drop from 0.533 to 0.131;
2) for Chamfer distance ($\downarrow$ is better), IF-Net results deteriorate from 0.134 to 0.444 and PIFuHD results worsen from 0.214 to 0.702;
3) for Normal consistency ($\uparrow$ is better), IF-Net results drop from 0.828 to 0.715 while PIFuHD results drop from 0.734 to 0.543.
Summarizing the three observations, we find OPlanes to be more robust to partial visibility.

\begin{table*}[h]
\renewcommand{\arraystretch}{1.0}
\begin{adjustwidth}{0.0cm}{}
\renewcommand\theadfont{}
\centering
\captionsetup{width=\linewidth}
\caption{\textbf{Performance across various visibility levels}.
For each cell, we report in the format of OPlane / IF-Net / PIFuHD.
Each OPlane result is averaged over three runs with different seeds and is reported in the format of $\texttt{mean}{\scriptsize \pm \texttt{std}}$.
The column ``Visibility Range'' refers to the range of the visibility percentage of a mesh. The higher the more visible.
\#Data denotes the number of evaluation entries in the corresponding range (row of the table).
We report the overall performance in the $5^\text{th}$ row while the $1^\text{st}$ to $4^\text{th}$ rows provide more fine-grained results.
Note, the $4^\text{th}$ row presents results for fully-visible objects.
As expected, when the visibility drops, the  performance drops too.
}
\label{tab: supp vis level}
\vspace{-0.2cm}
\begin{adjustbox}{width=\textwidth,center}
\setlength{\tabcolsep}{3pt}
{
\begin{tabular}{lcccccrrr} 
\toprule
 & \makecell{Partial\\Visibility} & \makecell{Visibility\\Range} & \#Data & IoU$\uparrow$  & Cham-$\mathcal{L}_1 \downarrow$ & \makecell{Normal\\Consistency} $\uparrow$ \\
\midrule
1 & Low & [0.069, 0.379) & 64 & 0.668{\scriptsize$\pm$0.021} / 0.365 / 0.131 & 0.289{\scriptsize$\pm$0.023} / 0.444 / 0.702 & 0.703{\scriptsize$\pm$0.008} / 0.715 / 0.543 \\
2 & Middle & [0.379, 0.690) & 552 & 0.618{\scriptsize$\pm$0.013} / 0.376 / 0.172 & 0.291{\scriptsize$\pm$0.012} / 0.461 / 0.654  & 0.710{\scriptsize$\pm$0.006} / 0.731 / 0.559 \\
3 & High & [0.690, 1.000) & 2167 & 0.699{\scriptsize$\pm$0.013} / 0.602 / 0.429 & 0.149{\scriptsize$\pm$0.008} / 0.204 / 0.322 & 0.753{\scriptsize$\pm$0.005} / 0.805 / 0.672 \\
4 & Full & [1.000, 1.000] & 1517 & 0.707{\scriptsize$\pm$0.013} / 0.644 / 0.533  & 0.109{\scriptsize$\pm$0.006} / 0.134 / 0.214 & 0.759{\scriptsize$\pm$0.005} / 0.828 / 0.734 \\
\midrule
5 & - & [0.069, 1.000] & 4300 & 0.691{\scriptsize$\pm$0.013} / 0.584 / 0.428  & 0.155{\scriptsize$\pm$0.008} / 0.216 / 0.332 & 0.749{\scriptsize$\pm$0.005} / 0.802 / 0.677 \\
\toprule
\end{tabular}
}
\end{adjustbox}
\end{adjustwidth}
\vspace{-0.5cm}
\end{table*}

\end{document}